\def\year{2019}\relax
\documentclass[letterpaper]{article} 
\usepackage{aaai19}  
\usepackage{times}  
\usepackage{helvet}  
\usepackage{courier}  
\usepackage{url}  
\usepackage{graphicx}  
\usepackage{amsthm}
\usepackage{booktabs}
\usepackage[utf8]{inputenc}

\newcommand{\citet}[1]{\citeauthor{#1} \shortcite{#1}}
\newcommand{\citep}{\cite}

\title{Cost-Based Goal Recognition Meets Deep Learning\thanks{An earlier version of this paper was published in PAIR (AAAI 2019 workshop).}}
\author{Mariane Maynard \and Thibault Duhamel \and Froduald Kabanza\\
PLANIART Laboratory\\
Université de Sherbrooke\\
Sherbrooke, Québec (Canada) J1K 2R1\\
\{mariane.maynard, thibault.duhamel, froduald.kabanza\}@usherbrooke.ca\\
}
\theoremstyle{definition}
\newtheorem{definition}{Definition}
\begin{document}
\maketitle
\begin{abstract}

The ability to observe the effects of actions performed by others and to infer their intent, most likely goals, or course of action, is known as a plan or intention recognition cognitive capability and has long been one of the fundamental research challenges in AI. Deep learning has recently been making significant inroads on various pattern recognition problems, except for intention recognition. While extensively explored since the seventies, the problem remains unsolved for most interesting cases in various areas, ranging from natural language understanding to human behavior understanding based on video feeds. This paper compares symbolic inverse planning, one of the most investigated approaches to goal recognition, to deep learning using CNN and LTSM neural network architectures, on five synthetic benchmarks often used in the literature. The results show that the deep learning approach achieves better goal-prediction accuracy and timeliness than the symbolic cost-based plan recognizer in these domains. Although preliminary, these results point to interesting future research avenues.
\end{abstract}

\section{Introduction}

The ability to infer the intention of others, also known as goal, plan, or activity recognition, is central to human cognition and presents a wide range of application opportunities in many areas. Human behavior is often the result of conscious and unconscious cognitive planning processes~\cite{schmidt_etal_78,baker_etal_09}. Therefore, to infer the intention of other people interacting with us, our brain is somehow able to predict what might be their goals or plans based on observations of clues from their actions. This capability is central to interact smoothly with people, to avoid danger in many situations, and to understand situations unfolding before us, such as predicting the behaviors of pedestrians when driving. Not surprisingly, there is intense research on intention recognition on many AI problems ranging from natural language understanding~\cite{wen_etal_17} and human-machine interaction~\cite{chen_etal_19} to autonomous vehicles~\cite{volz_etal_16} and security monitoring. 

Intention recognition is part of the more general problem of pattern recognition, with the critical nuance that it deals with goal-oriented patterns. Deep learning has been making significant inroads in recognizing patterns in general. Latest computer vision algorithms are now able to identify simple human behaviors involving short sequences of actions from videos, such as talking, drumming, skydiving, walking, and so on~\cite{simonyan_zisserman_14,hou_etal_18,yan_etal_16}. However, recognizing behaviors involving longer goal-oriented sequences of actions and produced by elaborate planning processes is another challenge yet barely tackled by end-to-end deep learning solutions~\cite{min_etal_14,min_etal_16,amado_etal_18}.

For a long time, various symbolic inference paradigms have been experimented to try to infer the intention from observations based upon handcrafted models, using probabilistic inference frameworks such as HMM~\cite{bui_etal_02}, Dynamic Bayesian Networks~\cite{charniak_goldman_93}, Markov logic~\cite{sadilek_kautz_10}, probabilistic grammar parsing~\cite{geib_goldman_09}, cost-based goal recognition~\cite{ramirez_geffner_10,masters_sardina_19}, etc. These approaches require that human experts provide models of behaviors (e.g., domain theories or plan libraries~\cite{sukthankar_etal_14}), serving as input to inference engines. However, like vision, language understanding, and other perception tasks, intent recognition is difficult to express in a model, and this often results in a biased or utterly inaccurate definition of the domain for the inference engine. The appeal of representational learning is indeed the ability to extract modeling features, otherwise difficult to explain for an expert, from data.  

In this paper, we show that familiar deep neural network architectures, namely dense, convolutional, and LTSM networks, can perform well on intention recognition problems in navigation domains compared to symbolic cost-based goal recognition algorithms considered as state of the art on this problem~\cite{ramirez_geffner_10,masters_sardina_19}. In this domain, we study the case of an agent (the observee) navigating in an environment, for whom the map is known \textit{a priori}, where several points of interest are their potential destinations. It is a synthetic benchmark, with some simplifications, but is a step towards solutions that will work eventually in more realistic environments.

While preliminary, results show that deep learning gives better and quicker goal-prediction accuracy than the state-of-the-art symbolic method. Comparisons on other academic benchmarks often used to evaluate symbolic plan recognizers also suggest that deep neural networks offer competitive performance. It seems that even a simple dense structure can learn abstractions underlying sequential decisions conveyed in the observed patterns of a goal-directed agent enough to outperform a cost-based approach. Before these experiments, we expected the latter to perform better since it is inherently tailored to deal with consecutive decisions. These surprising results raise exciting avenues of investigation that we discuss in the paper.

The rest of the paper follows with a brief review of the most related work, background, experiment methodology, experiment results, and conclusion.


\section{Related Work}

A few approaches combine deep learning and symbolic inference in different ways. For example,~\citet{granada_etal_17} use a deep neural network to recognize individual actions of an actor cooking recipes in a kitchen, and then use a symbolic algorithm, SBR, to infer the goal underlying an observed sequence of actions. This approach also requires a handcrafted model (plan library) representing abstractions of potential plans the agent could execute. Moreover, no mechanisms are allowing the handcrafted plan library to adapt to the classification errors made by the neural network recognizing individual actions.

The procedure in~\citet{bisson_etal_15} also makes use of a symbolic algorithm, which requires as input a sequence of observations of actions performed by an agent and a plan library. One component of the plan library representation is a probabilistic model of the choices the observed agent could make when selecting and executing plans from the plan library. A neural network learns this probabilistic model, whereas the rest of the plan library is handcrafted.

In both approaches, a symbolic inference engine makes the goal or plan predictions, not a neural network. Deep learning is involved only as an auxiliary procedure either to scan individual actions~\cite{granada_etal_17}, or to learn a probabilistic model~\cite{bisson_etal_15}. In contrast, in the experiments we discuss herein, a neural network makes all the inference.

To the best of our knowledge, \citet{min_etal_14} are among the first to use a goal recognition pipeline only made of a neural network. They use feed-forward n-gram models to learn the player's objective from a sequence of his actions in the \textsc{Crystal Island} game. The follow-up method in~\citet{min_etal_16} uses Long Short-Term Memory (LSTM) networks, better suited to learn patterns in sequences. In both approaches, the features fed to the neural network were engineered instead of merely being raw player's events such as mouse clicks and key presses. While these methods demonstrate favorable results in a specific domain, they do not include a systematic comparison to symbolic ones.

\citet{amado_etal_18} more recently introduced a deep learning pipeline to recognize the goal achieved by a player in different simple games (such as 8-puzzle and tower of Hanoi) from raw images, divided into three steps. First, they convert inputs into a latent space (which is a representation of state features) using a previous auto-encoder algorithm~\cite{asai_fukunaga_18}. Its properties are built to be reminiscent of a PDDL state representation. Then, an LSTM network utilizes it to perform a regression task, which is making a goal prediction in the latent space. Finally, the decoder reconstructs the goal image from its representation. While this approach does perform well on simple task-planning problems, it may not be applicable in real-life settings. The method indeed tries to extract an approximate domain structure (states representation reminiscent of a PDDL) from temporal changes in observation sequences, and it is unsure whether or not real data can be exploited to frame such rules.

Although some papers started to investigate deep learning for goal recognition, we are not aware of any systematic comparison between an end-to-end deep-learning pipeline and a symbolic or hybrid approach (in particular, directly and only on raw observations, which is the experiment specifically discussed herein).

\section{Background}

To understand the methodology used for the experiments, we first present some background on deep neural networks and cost-based goal recognition.

\subsection{The Problem}

The goal recognition problem consists in inferring the goal pursued by an actor from an observed sequence of action effects (and sometimes extract the plan pursued by the actor from these, extending the concept to plan recognition)~\cite{schmidt_etal_78}. There is a close link between goals, plans, and intentions. A plan is a sequence of actions achieving a goal, whereas an intention is a commitment to executing a plan. In general, one can infer a goal from a plan and vice-versa. Thus, in the AI literature, plan recognition has come to encompass all problems related to understanding goal-oriented behaviors, whether the focus is on inferring the goal, inferring intention, predicting the plan, or combinations of those three. 

The experiments discussed herein deal with inferring the distribution probability of goals by observing action effects. Given a sequence of observations $o_{\pi}=o_1, \ldots, o_n$, -- that may come directly from sensors or followed by relative prior parsing and processing -- and a set $G$ of potential goals that the agent might pursue, the problem is to infer a posterior probability distribution across $G$, $P(G|o_{\pi})$, representing the probabilities that the agent might be pursuing a goal given the observations. Note that a goal recognition problem is also a pattern recognition problem, but not vice-versa. That is, not all pattern recognition algorithms harness goal-directed behaviors, let alone, towards inferring the goals underlying goal-directed behaviors. 

\subsection{Deep Learning}

It is easy to cast a goal recognition problem as a supervised deep-learning problem. Given a set of sequences of observations ${\cal O}$ and a set of potential goals $G$, let us assume that there exists a true recognition function $f$ that maps perfectly each $o_\pi \in {\cal O}$ to its true goal $g_{o_\pi} \in G$, that is, $f(o_\pi) = g_{o_\pi}$.

While $f$ is unknown (this is what we want to infer), we assume we have access to a training dataset of paired examples $(o_\pi, g_{o_\pi})$ (we know the real goal $g_{o_\pi}$ for some $o_\pi \in {\cal O}$). A supervised learning algorithm will seek to approximate $f$ with a function $f'$ parameterized by some set of parameters $\theta$ that minimizes the number of erred predictions in our dataset of examples. In other words, $f'$ minimizes:

\[L=\sum_{n=0}^N l(f'(o_\pi^n; \theta), g_{o_\pi^n})\]

where $l$ is a loss function that is $0$ when $f'$ predicts accurately, and $> 0$ otherwise.

A single-layer neural network uses a simple linear transformation of the input using weight and bias parameters followed by a non-linear function in place of $f'$:

\[f'(o_\pi) = \gamma(Wo_\pi + b)\]

where $W$ and $b$ are the weight and bias parameters, respectively, and $\gamma$ is a non-linear function such as sigmoid, hyperbolic tangent (tanh), linear rectifier units (ReLU), or softmax. A (deep) neural network is a composition of several of these transformations, usually with a different set of parameters at each layer~\cite{goodfellow_etal_16}. These parameters are trained to minimize the loss function with the gradient descent algorithm.



There exist specialized types of networks that process data differently and are more fit for some forms of input and problems. For instance, convolutional neural networks (CNNs) use filters of parameters and the convolution operation to process 2D input, such as images or spatial information. Recurrent neural networks (RNNs) can memorize an internal state and process sequences, such as observed actions, making them better adapted to analyze dynamic behaviors than simple feed-forward architectures are. Long Short-Term Memory networks (LSTM) used by~\citet{min_etal_16} are types of RNNs that allow for better gradient propagation and thus show better learning results than vanilla RNNs on longer sequences.






\subsection{Symbolic Cost-Based Goal Recognition}

The intuition behind cost-based goal recognition is the \textit{principle of rationality}: people tend to act optimally to the best of their knowledge~\cite{baker_etal_09} and motor skills. Thus, one could infer the goal of an observed agent by trying to reason from their point of view, that is, trying to invert his planning process. It does not mean that we need to know his planning process.

\begin{figure}[htb]
    \centering
    \includegraphics[width=0.5\linewidth]{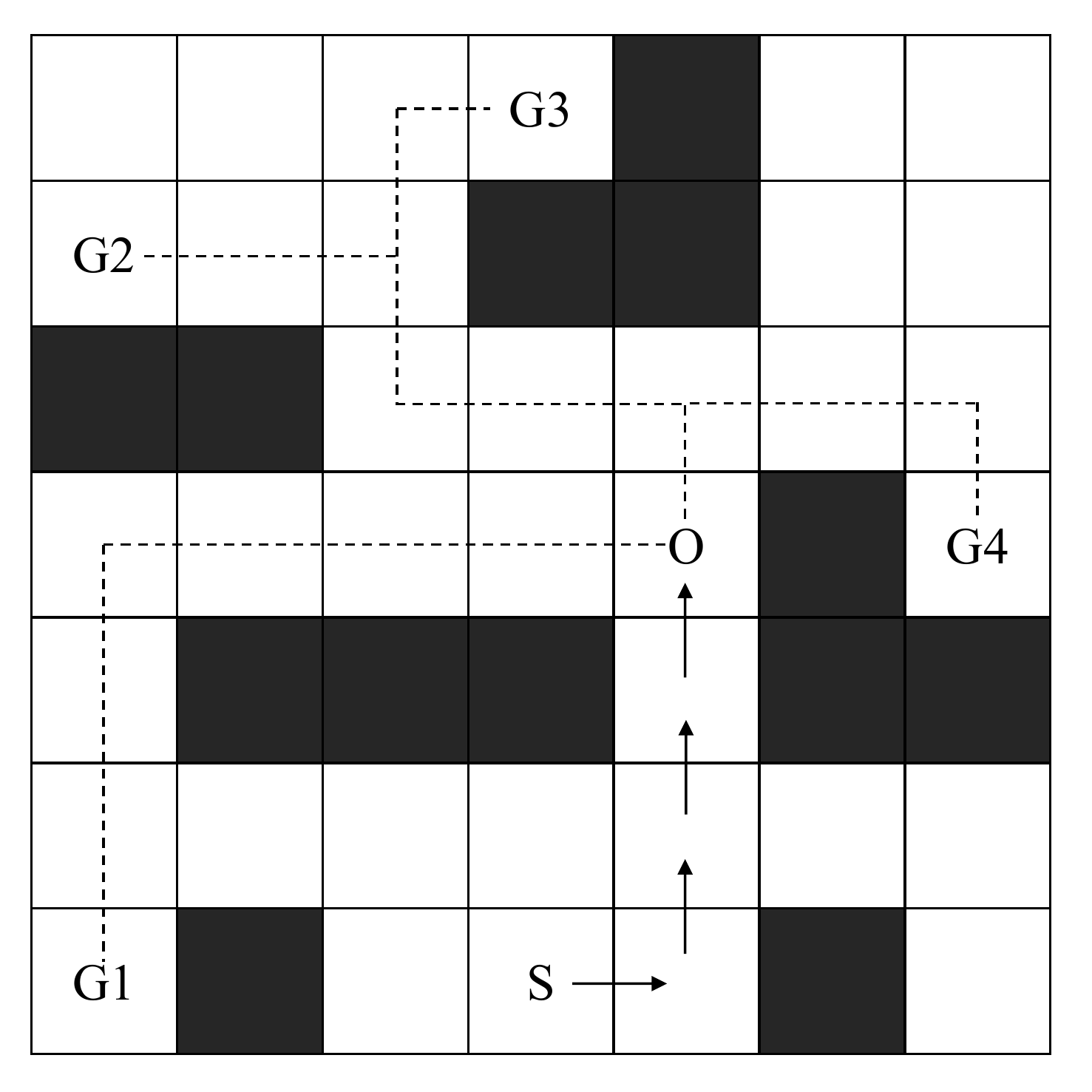}
    \caption{A navigation grid example, where the agent is constrained with obstacles.}
    \label{fig:gridexample}
\end{figure}

As noted by~\citet{ramirez_geffner_10}, given a sequence of observations, we could infer the probability that a given goal is the one being pursued by an agent by evaluating if his behavior observed so far is economical and might indeed commit to reaching that goal. To illustrate, consider the map in figure~\ref{fig:gridexample}, representing areas of interest (goals) $G_1, \ldots, G_4$, obstacles, and a sequence of observations of an agent moving around, starting from position $S$. From the observation so far $o_\pi=o_1\rightarrow\ldots\rightarrow o_4$, the agent logical goal is unlikely $G_1$, since we can find a shorter path from its start state to $G_1$ than the one they are currently taking. Intuitively, we can derive the likelihood of a goal by comparing the cost of an optimal plan consistent with the observations and the cost of an optimal plan not considering the observations. The higher the difference between these two costs is, the less likely the goal is. Formally, \citet{ramirez_geffner_10} calculate the likelihood of an observation sequence ${\cal O}$ to reach a goal $g$ as:

\[P(o_\pi|g) = \frac{e^{-\beta \Delta(s,g,o_\pi)}}{1 + e^{-\beta \Delta(s,g,o_\pi)}}\]

where $\beta$ is a positive constant determining how optimal we assess the observed agent's behavior to be. $\Delta$ is defined to be:

\[\Delta(s, g, o_\pi) = c(s, g, o_\pi) - c(s, g, \lnot o_\pi)\]

where $c(s, g, o_\pi)$ is the cost of the optimal plan $\pi_o$ between $s$ and $g$ that complies with the observations (all observed actions of $o_\pi$ are embedded monotonically in the plan) and $c(s, g, \lnot o_\pi)$ is the cost of the optimal plan $\pi_{\lnot o}$ that does not comply with the observations ($\pi$ does not embed $o_\pi$).

From $P(o_\pi|g)$, we can derive the posterior probability of the goal using the Bayes rule: $P(g|o_\pi) = \alpha P(o_\pi|g) P(g) \forall g \in G$, where $P(g)$ is the prior probability (often assumed to be uniform) and $\alpha$ is a normalization factor.

In principle, a planner can be used to compute plan costs~\cite{ramirez_geffner_10}. However, calculating a plan, even in the simple case of a deterministic environment under full observability, is NP-Complete~\cite{cooper_90}. It is not realistic in situations where an agent needs to infer the intention of others quickly. Approximate plan costs, computed by suboptimal planners that run faster than optimal ones, can be used to deduce approximate distribution~\cite{ramirez_geffner_09}. They can be helpful in situations where the most important thing is to identify the most likely goals. Nonetheless, even heuristic planners that compute suboptimal plans still take too much time for most real-time applications.

We can avoid some calls to the planners by incorporating heuristic functions directly into the inference process. \citet{vered_kaminka_17} introduced such heuristics that judge whether a new observation may change the ranking of goals and whether to prune a goal or not. However, they become useless in more complex problems where the goals cannot be pruned early and do not reduce the number of calls to the planner.

A practical approach to cost-based goal recognition is to compute the plan costs offline. This way, instead of invoking a planner, we have a lookup in a table or a map of plan costs. For navigation problems, where the issue is to predict the destination of an agent moving around, \citet{masters_sardina_19} describe an approach for accurately pre-computing plan costs by relaxing \citet{ramirez_geffner_10}'s algorithm with -- practically -- no loss in accuracy. It is overall the same, but they compute the cost difference to instead be $\Delta(s, g, n) = c(n, g) - c(s, g)$ where $n$ corresponds to the last seen position of the observed agent. This relaxation not depending on the whole observation sequence avoid computing as many different costs as needed by \citet{ramirez_geffner_10}, making them easier to be stored beforehand. However, it is quite limited in application to the -- discrete -- navigation domain.

In general, however, there is no well-known method of accurately pre-computing and storing plan costs for all possible combinations of initial and goal states for an arbitrary domain. \citet{sohrabi_etal_16} compute the top-k plans for each goal and calculate the goal inference by summing the probability of plans in the set achieving this goal, where the likelihood of a plan does not only depend on its cost but also to what degree it complies to the observations. The problem is that the required number of plans is high (1000) to have results comparable to \citet{ramirez_geffner_10}'s. Other various recent studies present different ideas to reduce planners' compute time. For instance, \citet{e-martin_etal_15} calculate cost interaction estimates in plan graphs, while \citet{pereira_etal_17} use landmarks, with the idea that goals with a higher completion ratio are the likely ones. However, their solutions are less accurate since they are mere approximations of plans generated by an optimal planner.

\section{Comparison Methodology}

To compare cost-based goal recognition to deep learning, we used five synthetic domains often selected to evaluate the performance of a symbolic plan recognizer, as referenced above. Ultimately, we want to examine plan recognizers using real-world benchmarks. Meanwhile, the synthetic domains can provide some useful insight.


\begin{enumerate}
\item \textsc{Navigation}: Predicting the goal destination of an agent navigating a map~\cite{masters_sardina_17}. The domain consists of 20 maps from StarCraft, provided by MovingAI\footnote{MovingAI Lab: https://movingai.com/}, downscaled to 64x64 pixels, where the agent can perform actions limited to the first four cardinal directions. We generated the goal recognition problems by placing one initial position and five goals on the maps.

\item \textsc{Intrusion Detection}: Predicting the goals of network hackers with their activities~\cite{geib_goldman_02}. The observed agent is a user who may perform attacks on ten hosts. There are six possible goals that the hacker might reach by performing nine actions on those servers. Observation sequences are typically between 8 and 14 observations long.

\item \textsc{Kitchen}: Inferring the activity of a cook in a smart home kitchen~\cite{wu_etal_07}. The cook can either prepare breakfast, lunch, or dinner (possible goals)~\cite{wu_etal_07}. He may manipulate objects, use them, and perform numerous high-level activities. Observation sequences are typically between 3 and 8 actions long.

\item \textsc{BlocksWorld}: Predicting the goal of an agent assembling eight blocks labeled with letters, arranged randomly at the beginning~\cite{ramirez_geffner_09}. Achieving a goal consists in ordering blocks into a single tower to spell one of the 21 possible words by the use of 4 actions. Observation sequences are typically between 6 and 10 actions long. 

\item \textsc{Logistics}: Predicting package delivery in a transport domain. Six packages must be conveyed between 6 locations in 2 different cities, using one airplane, two airports, and two trucks~\cite{ramirez_geffner_09}. There are six possible actions available to achieve ten distinct goals. Observation sequences are typically between 16 and 22 actions long.

\end{enumerate}

The observation data for the four last benchmarks are available at https://github.com/pucrs-automated-planning/goal-plan-recognition-dataset.

\begin{figure*}[htb]
    \centering
    \includegraphics[width=0.95\textwidth]{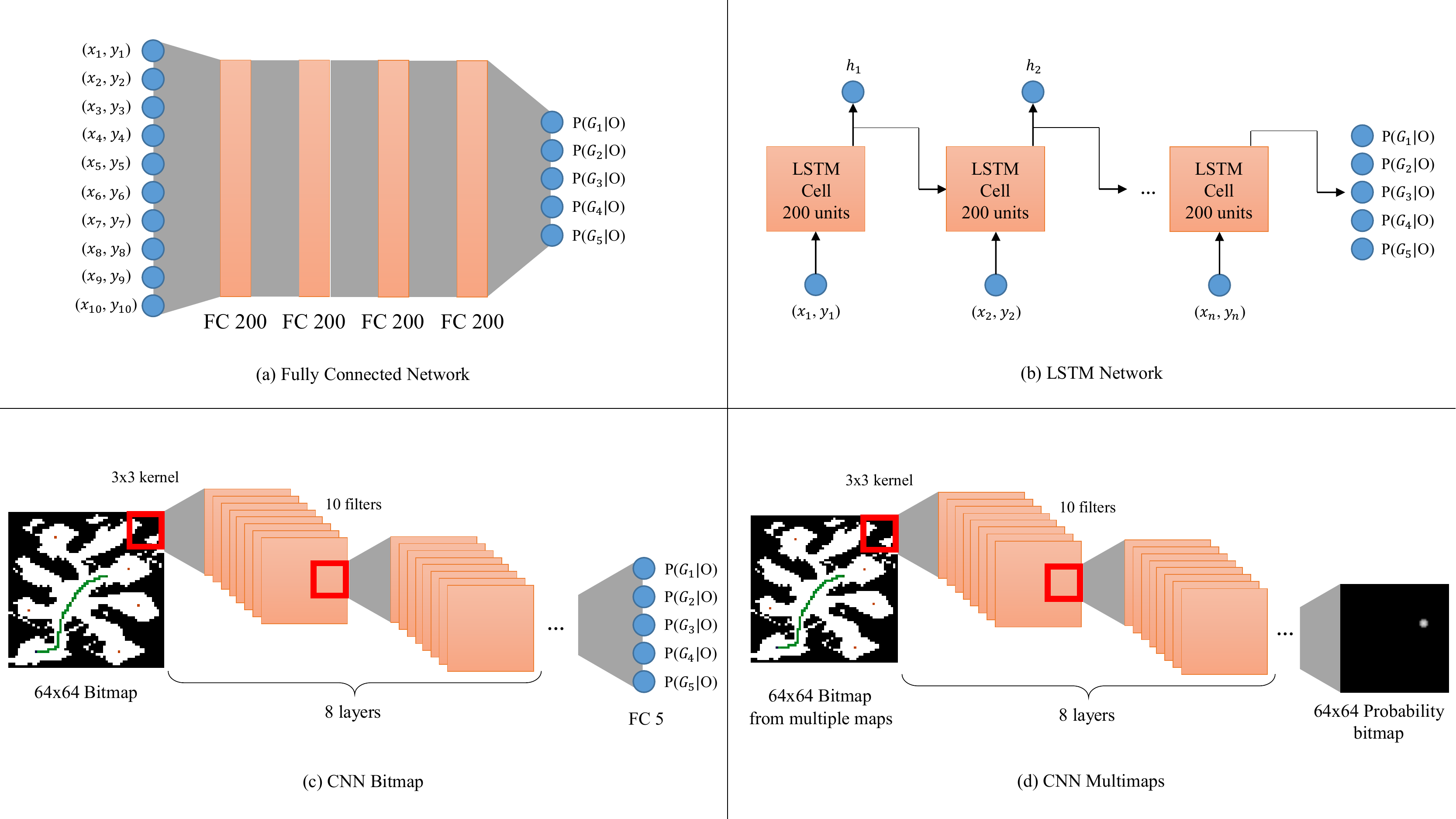}
    \caption{Representation of our architectures for the navigation domain. $(x_i,y_i)$ stands for the coordinates of the agent's location in the grid. $(a)$, $(b)$, and $(c)$ were trained on a single map, while $(d)$ was trained on multiple maps.}
    \label{fig:Networks}
\end{figure*}

For the navigation benchmark, we used four different neural network architectures (see figure~\ref{fig:Networks}): a fully connected network (FC), an LSTM network, and two convolutional neural networks (CNN). We felt both the LSTM and CNN appropriate for this domain, given that the former usually performs well learning from sequences, whereas the latter is suitable to learn from spatial data (maps in our case). 

We trained the first three networks on problems generated from a single map. We additionally trained a convolutional network (CNNMultimaps) on multiple ones, regardless of their goals, start and obstacle positions, to see if and how it could generalize across multiple navigation domains.

Here is a thorough description of the network architectures:

\begin{enumerate}
\item FC: this network contains four dense layers of 200 units and one output layer of 5 units representing the goal probability distribution.
\item LSTM: this network as a single LSTM layer of 200 units and a dense output layer of 5 units.
\item CNN (CNNBitmap): this network has eight convolutional layers of 10 filters of size 3x3, respectively. The resulting features are flattened and passed to a dense layer of 5 units.
\item CNNMultimaps: the first eight layers of this network are the same as in the CNNBitmap, followed by an additional convolution layer of one 3x3 filter instead of a dense layer.
\end{enumerate}

Since we trained and tested the methods FC, LSTM, and CNNBitmap on the same map, with goals identified in advance, it was possible to deduce a probability distribution array of fixed size (five here). However, we could not make this assumption for the general fully convolutional method (CNNMultimaps) trained on multiple, different maps, which instead outputs a probability distribution over the entire grid, representing a spatial belief about the agent's goal, allowing any number of goals and positions in general.

For the four other domains, we used a fully connected network with three dense layers of 256, 32, and 5 units, respectively. We compare it with original \citet{ramirez_geffner_10}'s method, since there is yet no proven method for pre-computing plan costs -- or approximations of them -- for these domains without a significant loss in accuracy~\cite{e-martin_etal_15,pereira_etal_17,vered_kaminka_17}.


Besides the architecture, implementing neural networks involves the choice of specific parameters, activation functions, and optimization algorithm.  Given that we want to find a correct goal amongst a set of possible ones and work with probabilistic scores, we quantify the loss with the categorical cross-entropy function and work with the accuracy metric, which is the percentage of correct predictions. A prediction is said to be correct if its highest output probability corresponds to the true goal. In case of ties, we consider a random uniform draw between all the goals having the same top probability. In cost-based goal recognition literature, alternative accuracy metrics are often used, such as metrics using a threshold~\cite{pereira_etal_17,sohrabi_etal_16}, or simply an accuracy metric where ties are not randomly disambiguated and instead considered as an accurate prediction \cite{ramirez_geffner_10,e-martin_etal_15,sohrabi_etal_16}. However, we find them highly artificial and unfit to evaluations of real-world applications, so we chose to consider the top 1 only, which should account for lower accuracy values. It is also important to note that we apply the same metric to every method.

Hidden layers are activated with the ReLU function, while the output layer is activated with the softmax function. To train the networks, the Adam optimizer \cite{kingma_etal_14} is used, with a learning rate of 0.001, $\beta1$ of 0.9, $\beta2$ of 0.999 and no decay. To prevent overfitting, we also used dropout \cite{srivastava_etal_14} for all layers with a drop chance set to 0.1 or 0.2. Finally, inputs were shuffled uniformly before training.

\section{Experiments and Results}

We present the experiments and discuss their results in this section, including complete details about the training and test datasets. For all domains, the datasets are split 80\%-20\% for training and test.

\subsection{Navigation Domain}

As mentioned above, we trained four networks for the navigation benchmark. The first three (FC, LSTM, CNNBitmap) were trained for 15 epochs on observations from a single map, with 100 observed paths. We also trained CNNMultimaps on all the available maps for 100 epochs. To mimic suboptimal behavior, we started by generating noisy optimal paths to these goals with a modified A* algorithm, using what we define as an $\epsilon$-over-estimating heuristic:
\begin{definition}
An \textit{$\epsilon$-over-estimating heuristic} is a function that returns an admissible quantity $h'$ with a chance of $1-\epsilon$, and $h' + \delta$ otherwise, where $\epsilon \in$ [0, 1] and $\delta > 0$.
\label{def:epsheuristic}
\end{definition}

We truncated the generated paths to measure how our networks could handle early predictions in an online application: both training and test sets consist of partial or complete sequences of observations truncated at the first 25\%, 50\%, 75\% and 100\% of the sequence, such that we can evaluate performances for partial as well as complete observability. It is important to note that this notion of partial observability differs from the usual literature definition: in many papers \cite{ramirez_geffner_10,pereira_etal_17,e-martin_etal_15,sohrabi_etal_16}, a certain percentage of observations is missing, but across the \textit{whole} sequence. In opposition to that, to mimic real-time predictions, we cut the observation sequences to a given percentage, and drop every following observation. We estimate that this idea of early observability is more realistic as it enables online resolution of goal recognition problems.

We used $(x, y)$ coordinates as input for the FC network and LSTM methods. As paths lengths may differ, we eventually retained a fixed number of positions among the ones available to form inputs of fixed size, padding shorter sequences with zeros. We fed 4-channel bitmaps to both CNNs, where each channel embeds information about either the initial position, the potential goals, the observations, and walkable locations that are neither of the above.

For \citet{masters_sardina_17}'s method (labeled M-S), we only considered the last position of the sub-paths. Cost maps were generated using optimal paths returned by the A* algorithm and stored offline. To compute the posterior probabilities, we assumed prior probabilities to be uniform and used a value of 1 for the $\beta$ parameter.

\begin{figure}[htb]
    \centering
    \includegraphics[width=\linewidth]{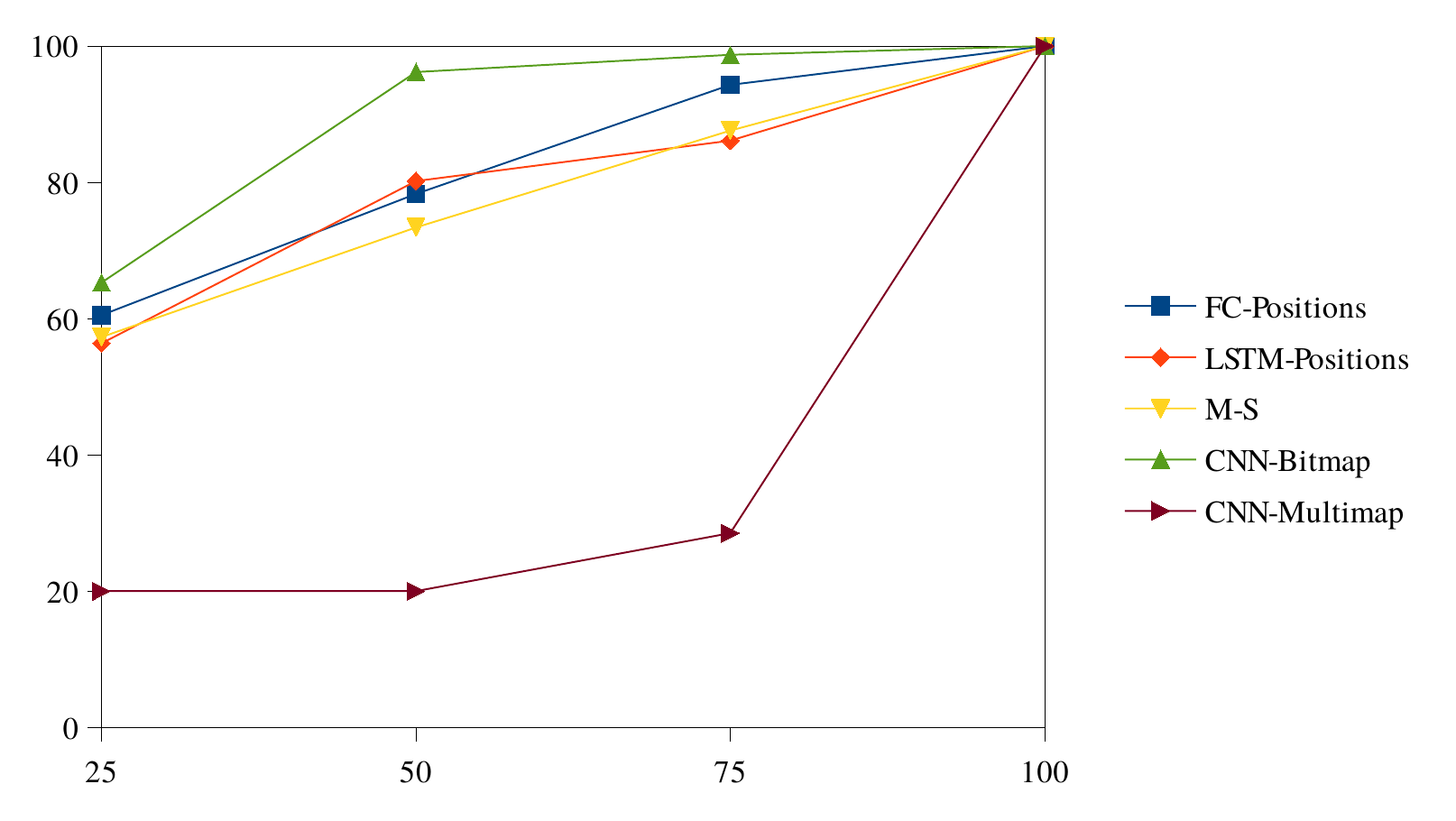}
    \caption{Results of accuracy depending on the percentage retained from the complete observed path, in the navigation domains.}
    \label{fig:graph1}
\end{figure}

We compared the accuracy of those four different networks on test sets with M-S. Results are shown in figure~\ref{fig:graph1}. The Y-axis represents the average accuracy on ten different maps. The X-axis refers to the percentage sampled from total paths in the test set. 

As can be seen, method CNNBitmap ranks first. The reason could be that the convolution filters of the network help reason about the 2D structure of the grid and the observed path, as expected. FC and LSTM methods perform well too, but it seems that learning from coordinates is more complicated, or more imprecise, that learning directly from bitmaps in such a navigation domain. 

Surprisingly, M-S was outperformed at least by CNNBitmap and FC. The reason might be that generated A* tracks stayed somehow deterministic despite the noisy behavior, and thus, even in the case where multiple optimal paths to a goal exist, similar routes were always chosen for that goal. The neural networks thus quickly learned to fit these specific paths, even though earlier subsets could go to either goal. This bias in the data incorporated by the generation process could be problematic, but we argue otherwise. In real-world applications involving human agents, people usually take the same road even when multiple ones that are as good -- or even better -- exist. Data is therefore not uniformly distributed between every candidate road. The capacity of neural networks to learn this bias and adjust for particular contexts and individuals is one of the properties that makes them appropriate for goal recognition in real-life applications. Additionally, in the case of cost-based algorithms, even though all available data is used to compute costs, the final prediction is only achieved based on them, which represents a gradual loss of information.

The convolutional network trained and tested on all maps (CNNMultimaps) shows relatively incorrect early predictions (20\% accuracy for five goals is just a random prediction), proving there is still room for improvement to generalize to multiple maps. Nonetheless, the method can already create a link between a complete path and a goal (that is, learning but not predicting), and we may significantly improve its results using specialized architectures, such as value iteration network~\cite{tamar_etal_17} and visual relational reasoning~\cite{watters_etal_17}. We are currently working on improving its results.

Computing plan costs takes time, even offline. The results suggest that training neural networks, even if computationally complex, may be advantageous in this regard thanks to the trivially parallelizable nature of its operations and the computation power of modern hardware. However, a computation time comparison does not enlighten new advantages for this kind of context. Table~\ref{tab:times1} gives a summary of offline and online computation times. The LSTM networks have longer training times but may generalize better to longer sequences of observations with bigger sliding windows (since we fixed the maximum number of observations input to 10 and thus do not benefit sufficiently from LSTM's training power over sequences). The CNN trained on multiple maps takes a long time to train but could have the potential to generalize to every navigation problem so that it would require no additional training for unseen configurations. Symbolic approaches have no need for training nor dataset, but knowledge about the domain is needed to handcraft the model, and costs must be generated for every new map, whether it is offline or online (during prediction).

\begin{table}[htb]
    \centering
    \begin{tabular}{@{}l|c|c@{}}
    \toprule
                                    & T             & P                \\
    \midrule
    FC & 10 s          & $10 \mu$s        \\
    LSTM                    & 30 s          & 4 ms             \\
    CNNBitmap               & 10 s          & 4 ms             \\
    CNNMultimaps      & 20 min        & 4 ms             \\
    R-G             & 0             & 1 s              \\
    M-S     & 7 s           & $10 \mu$s        \\
    \bottomrule
    \end{tabular}
    \caption{Comparison of rough average computation times of the evaluated approaches on the navigation domain. T is the offline computation time, while P is the online prediction time.}
    \label{tab:times1}
\end{table}

\subsection{Other Domains}

The navigation benchmark deals with path-planning problems requiring much less knowledge than the other four domains. Those last benchmarks correspond to task-planning problems, involving constraints that differ from those in the navigation benchmark, thus requiring different kinds of domain representations (represented using the Planning Domain Definition Language (PDDL) as in~\citet{ramirez_geffner_10}).

We trained a fully connected network during 15 epochs, with 1000 to 3000 examples depending on each domain. We also trained an LSTM on these examples, but it ended up taking more time without providing significant result improvements.

A training example in the datasets is a sequence of observations from PDDL files. Each observation in the sequence is one action type plus its arguments, both transformed into a one-hot vector. The neural network receives the complete sequence of transformed observations. To match a fixed input size, sequences shorter than the maximum size are padded with zeros and shifted $\mathit{maxSize}-\mathit{size}+1$ times (for instance, if one observation is $A B$ and the maximum size is 4, 3 new observations will be created: $A B 0 0$, $0 A B 0$, $0 0 A B$), hence generating new training data.

In the case of \citet{ramirez_geffner_10}'s method, labeled R-G, the costs were generated online, as first implemented by the authors, from optimal plans found by the HSP planner. The $\beta$ parameter value was one, and the prior probabilities of the goals were presumed to be uniform.

\begin{figure}[htb]
    \centering
    \includegraphics[width=\linewidth]{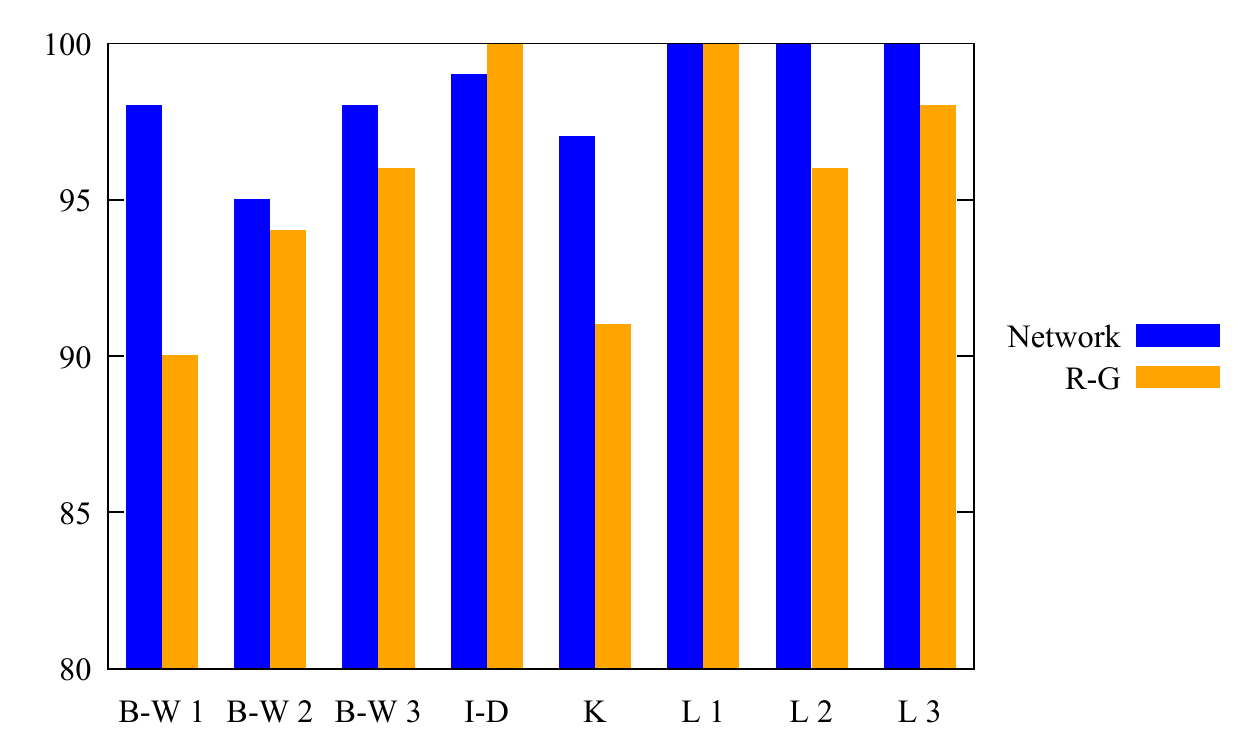}
    \caption{Results of accuracy for the task-planning domains (B-W, I-D, K, and L stand for \textsc{Blocks World}, \textsc{Intrusion Detection}, \textsc{Kitchen} and \textsc{Logistics} respectively).}
    \label{fig:graph2}
\end{figure}

Results in figure~\ref{fig:graph2} show the accuracy for both methods. The fully connected network outperforms the R-G approach almost every time. We provide a similar explanation for these results: generated sequences tend to be biased for each goal, and the network learned it. In addition to producing higher prediction rates, networks are also quicker: on such problems, the training part takes approximately one minute to infer reusable weights, and a prediction requires approximately 1ms. The R-G approach does not require training nor offline computation, but provides a prediction in minutes, sometimes hours, which is very long and cannot run for real-time decision making. A suboptimal planner might reduce computation times, but we can reasonably assume that it would remain above several minutes or so for each goal prediction.

\section{Conclusion}

Although still preliminary, these results suggest that deep learning outperforms symbolic inverse planning, at least in the five domains considered. We plan to pursue this experimentation in real-world settings where we can gather data, including video games. We also plan to try other deep neural networks~\cite{goodfellow_etal_16}, symbolic methods, multi-agent configurations, sensor limitations (partial observability vs. full observability), attitudes between the observed agent and the observer (cooperative, adversarial, neutral) and different domains of application. 

In some applications, the plan recognizer needs to explain the rationale of its inferences. To do so, extracting a meaningful explanation from a neural network remains a challenge. In contrast, the representation of symbolic plan recognizers directly answers to this question, except that, as we have argued, those approaches are difficult to ground in real-world environments. It suggests that the exploration of hybrid approaches, such as those discussed in the related section, remains worth pursuing.

\section{Acknowledgements}

The Natural Sciences and Engineering Research Council (NSERC) of Canada and the \textit{Fonds de recherche du Québec -- Nature et technologies} (FRQNT) supported the work with grants, and Compute Canada provided computing resources. The NVIDIA Corporation donated the Quadro P6000 used for this research. We are also thankful to Julien Filion, Simon Chamberland, and anonymous reviewers for their insightful feedback that helped improve the paper.

\bibliographystyle{aaai}
\bibliography{references}
\end{document}